%% file: iclr2020_conference.tex
\documentclass{article} 
\usepackage{iclr2020_conference,times}
\usepackage{wrapfig}
\usepackage{comment}

\input{math_commands.tex}

\usepackage{graphicx}
\usepackage{hyperref}
\usepackage{url}
\usepackage{authblk}
\usepackage{blindtext}

\def\rotatecharone#1{\rotatebox[origin=c]{180}{#1}}

\title{Assessing Human Translations from French to Bambara for Machine Learning: a Pilot Study}

\author[1]{Michael Leventhal}
\author[2]{\{Allahsera Tapo, Marcos Zampieri, and  Christopher M. Homan\}}
\author[3]{Sarah Luger}

\affil[1]{Centre National Collaboratif de l'Education en Robotique et en Intelligence Artificielle (RobotsMali) \textit{mleventhal@robotsmali.org}}
\affil[2]{Rochester Institute of Technology \textit{\{\{aat3261, marcos.zampieri\}@rit.edu, cmh@cs.rit.edu\}}}
\affil[3]{Orange Silicon Valley \textit{sarah.luger@orange.com}}

\iclrfinalcopy 
\begin{document}

	\maketitle
	
	\begin{abstract}
		We present novel methods for assessing the quality of human-translated aligned texts for learning machine translation models of under-resourced languages. Malian university students translated French texts, producing either written or oral translations to Bambara. Our results suggest that similar quality can be obtained from either written or spoken translations for certain kinds of texts. They also suggest specific instructions that human translators should be given in order to improve the quality of their work.
	\end{abstract}

	\section{Introduction}
	
	Mali has a literacy rate of 35\%, fifth-to-last in the world, which coincidentally is also its human development index ranking \citep{con2019hdr}. A survey by \cite{bleck2015education} revealed a strong correlation between use of French and the knowledge needed for informed citizenship and political engagement.
	Bambara, the vehicular language of Mali, is spoken by about 80\% of the population; French, the official language of Mali and the language of written communication, is spoken by an estimated 21\% \citep{lafage1993french}. Machine translation is a promising technology \citep{DBLP:journals/corr/abs-1906-05685} for dramatically increasing Bambara speakers' access to information that can impact education, health, and economic development.
	
	Bambara has a writing system that is taught in many primary schools, but its actual use outside the classroom remains rare compared to French. It is a severely under-resourced language for machine translation, as the quantity of bilingual aligned texts is extremely small. We explore the feasibility of crowdsourcing to create aligned translations from French to Bambara. Our primary source for translators will likely be Malian university students, as they are necessarily fluent in French, the language of instruction in Mali. However, many students have not learned to read or write Bambara. Thus, we assessed the quality of translations from written French to spoken Bambara and compared them to those from written French to written Bambara.
	
	\section{Methods}
	
	Seven translators participated, four using the written-written and three the written-oral method. The texts were: an excerpt of a Wikipedia article on a Mali-related topic (195 words, 10 sentences) and introductions to stories from Malian televised news broadcasts (428 words, 22 sentences).
	We chose a Bambara speaker that self-identified as having an advanced level of reading and writing knowledge to evaluate the translations. Having confirmed with the Académie Malienne des Langues, the official government agency for national languages, that no standard assessment of Bambara competence exists, we obtained BLEU scores for his translations of 3 sentences from a published bilingual medical text, using the text's Bambara translations as the reference. The evaluator obtained an average score of .40. The evaluator rated the translations qualitatively according to three criteria: the extent to which the translation accurately conveyed the information in the source text, the literal closeness of the Bambara translation to the source text, and the extent to which the translation used standard Bambara orthography (if written) and vocabulary (both written and oral). The evaluation was measured on a percentage scale where 0\% would be given to a translation that bore no relation to source text and 100\% to a translation that perfectly represented the source text according to the specific criteria. We also measured the translations' BLEU scores. 
	Since we lacked reference translations for these texts, we ranked our translations pairwise, with each one in turn serving as a reference to the others.
	
	We gave the translators minimal written guidelines in French, with additional oral guidance in Bambara and French. As we wanted to simulate conditions using a large number of unsupervised non-professional translators we instructed the participants to only make a single pass at the translation, not to use reference sources, and to make a best approximation in the event that they were unsure of a translation. Our experiment showed that more detailed guidelines are necessary and it also indicated various kinds of post-processing of the translations that would be helpful. For example, proper nouns were translated in many different ways, including being: kept in their French form, phonetically Bambara-ized with forms reflecting differences in pronunciation, and, occasionally, written using the standard Bambara spelling. Numbers were sometimes written as words, other times as numbers and sometimes as words followed by the numeric value in parentheses, a convention often used in Malian texts. Contractions and agglutinated words were sometimes contracted or agglutinated, sometimes not. It is likely that, under a more tightly controlled translation regime, the BLEU scores will be much higher.
	
	\section{Results}
	
	Table \ref{tab:summary} shows the results for the Malian news broadcast and the Wikipedia article.
	The differences in average scores between the news broadcast and the Wikipedia article, aside from the small sampling, most probably reflect the different challenges of the texts. The news broadcast is essentially an oral text and it is easier to reproduce the exact meaning with a more colloquial style. The Wikipedia article has long and complex sentences, making it easier to miss details and nuances while the translator hews closer to the French source and falls back on more formal standard Bambara.

	The translations of the news broadcast showed a relatively limited difference in meaning and use of standard Bambara between written and oral translations, but significant difference in the literalness of the translations. The relatively large standard deviations shown in Table \ref{tab:stddev} indicate a wide range of quality between translators and translations, suggesting that screening translations based on basic quality metrics may be necessary and effective.
	
	\begin{table}
		\begin{tabular}{r|rrrrrr}
			&Malian news && &Wikipedia&\\
			&Written&Oral&Overall&Written&Oral&Overall\\
			\hline
			Exact meaning & 0.830 & 0.770  & 0.840 & 0.870 & 0.530 & 0.730\\
			Literalness & 0.730 & 0.530 & 0.640 & 0.830 & 0.580& 0.760\\
			Standard Bambara  &  0.740 & 0.790 &  0.760 & 0.830 & 0.850 & 0.830\\Highest BLEU Pair & 0.408 & 0.363 & 0.408 & 0.645 & 0.377 & 0.645
		\end{tabular}
		\caption{Malian news broadcast and Wikipedia article translation ratings and BLEU scores.}
		\label{tab:summary}
	\end{table}
	
	\begin{table}
		\begin{tabular}{r|rrrrrr}
			&Malian news && &Wikipedia&\\
			&Written&Oral&Overall&Written&Oral&Overall\\
			\hline
			Exact meaning & 0.181 & 0.208  & 0.197 & 0.206 & 0.186 & 0.220\\
			Literalness & 0.130 & 0.298 & 0.243 & 0.238 & 0.211 & 0.244\\
			Standard Bambara  &  0.234 & 0.178 &  0.207 & 0.171 & 0.068 & 0.144\\
		\end{tabular}
		\caption{Score variance standard deviation}
		\label{tab:stddev}
	\end{table}

	Here is an example of a written and oral translations of French text, followed by the qualitative and quantitative evaluation of quality.

	French Source\\
	Objectif réfléchir à de nouvelles stratégies de lutte contre le terrorisme qui continue de faire des victimes dans le sahel.

	English\\
	Objective to reflect on new strategies to fight terrorism which continues to claim victims in the Sahel.

	Written Translation\\
	Laj$\varepsilon$ ni kun tun ye ka hakili jakab\rotatecharone{c} k$\varepsilon$ f$\varepsilon\varepsilon$r$\varepsilon$ kuraw la banbaanciw juguya la miniw b$\varepsilon$ ka ciy$\varepsilon$nni k$\varepsilon$ Saheli k\rotatecharone{c}n\rotatecharone{c}n\rotatecharone{c}n\rotatecharone{c}na la

	Oral Translation\\
	A kun tun ye ka miriya kuraw ta ka banbaanciw k$\varepsilon$l$\varepsilon$li sira kan o mun bi ka k$\varepsilon$ sababu ye ka fagali caman k$\varepsilon$ saheli k\rotatecharone{c}n\rotatecharone{c}nana

	The highest scoring BLEU pairs in all but one of the aligned translations from the news source were between oral and written translation methods. In the one remaining case written-written and written-oral pairs had approximately the same high BLEU scores, the scores being the highest from among all the news source translations.
	
	The translations of the Wikipedia article show that the meaning of the text was captured much better in the written-to-written translations. With only one exception, the highest scoring BLEU pairs were the written-to-written translations.
	These results suggest that written-to-written translation may be best for more complex texts while oral translations works well on simple texts.
	
	\section{Conclusion}
	
	Our experiments increased our confidence that oral translation using non-professional translators is a viable process for at least part of our effort to collect aligned texts for Bambara. Given that our pool of potential translators consists primarily of students that do not read or write Bambara, this lends support to the thesis that properly marshalling these resources can significantly contribute to a machine translation effort for our under-resourced, predominately oral language. Our tests suggest that initial efforts with oral translation might best focus on simpler French texts imbued with oral speech patterns. Having collected and compared a small sample of translations it was easy to identify a large number of semantically inconsequential differences in translation that might be normalized to produce better machine translation outcomes either through translator guidelines or by automated text cleaning.
	
	\bibliography{references}

	\bibliographystyle{iclr2020_conference}

\end{document}

%% file: math_commands.tex
\usepackage{amsmath,amsfonts,bm}

\def\eqref#1{equation~\ref{#1}}

\def\1{\bm{1}}

\DeclareMathAlphabet{\mathsfit}{\encodingdefault}{\sfdefault}{m}{sl}
\SetMathAlphabet{\mathsfit}{bold}{\encodingdefault}{\sfdefault}{bx}{n}

